\documentclass[runningheads]{llncs}
\usepackage{graphicx}

\usepackage{tikz}
\usepackage{comment} 
\usepackage{amsmath,amssymb} 
\usepackage{color}
\usepackage{subfigure}
\usepackage{pifont}
\usepackage[misc]{ifsym}

\usepackage{array}
\usepackage[hidelinks]{hyperref}
\usepackage{booktabs}
\usepackage{algorithm}
\usepackage{algorithmic}
\hypersetup{
	colorlinks=true,
	linkcolor=blue,
	filecolor=magenta,      
	urlcolor=cyan,
}
\usepackage{wrapfig,lipsum,booktabs}

\begin{document}
\pagestyle{headings}
\mainmatter
\def\ECCVSubNumber{100}  

\title{Geometry Constrained Weakly Supervised Object Localization} 

\titlerunning{Geometry Constrained Weakly Supervised Object Localization}
%
\author{Weizeng Lu\inst{1,2} \and
 Xi Jia\inst{3} \and
Weicheng Xie\inst{1,2} \and
Linlin Shen\inst{1,2}$^{(\textrm{\Letter})}$ \and
Yicong Zhou\inst{4} \and
Jinming Duan\inst{3}$^{(\textrm{\Letter})}$ }
\authorrunning{W. Lu et. al.}
%
\institute{Computer Vision Institute, School of Computer Science and Software Engineering, Shenzhen University, China\and
Shenzhen Institute of Artificial Intelligence and Robotics for Society, China\and
School of Computer Science, University of Birmingham, United Kingdom\and
Department of Computer and Information Science, University of Macau, China\\
\email{luweizeng2018@email.szu.edu.cn, x.jia.1@cs.bham.ac.uk, wcxie@szu.edu.cn, llshen@szu.edu.cn, yicongzhou@um.edu.mo, j.duan@cs.bham.ac.uk}}
\maketitle

\begin{abstract}
We propose a geometry constrained network, termed GC-Net, for weakly supervised object localization (WSOL). GC-Net consists of three modules: a detector, a generator and a classifier. The detector predicts the object location defined by a set of coefficients describing a geometric shape (i.e. ellipse or rectangle), which is geometrically constrained by the mask produced by the generator. The classifier takes the resulting masked images as input and performs two complementary classification tasks for the object and background. To make the mask more compact and more complete, we propose a novel multi-task loss function that takes into account area of the geometric shape, the categorical cross-entropy and the negative entropy. In contrast to previous approaches, GC-Net is trained end-to-end and predict object location without any post-processing (e.g. thresholding) that may require additional tuning. Extensive experiments on the CUB-200-2011 and ILSVRC2012 datasets show that GC-Net outperforms state-of-the-art methods by a large margin. Our source code 
is available at \textcolor[rgb]{1,0.5,0.9}{https://github.com/lwzeng/GC-Net}.
\end{abstract}

\section{Introduction}
In a supervised setting, convoluational neural network (CNN) has showed an unprecedented success in localizing objects under complicated scenes~\cite{girshick2014rich,redmon2016you,liu2016ssd}. However, such a success is relying on large-scale, manually annotated bounding boxes (bboxes), which are expensive to acquire and may not be always accessible. Recently, researchers start to shift their interests to weakly supervised object localization (WSOL)~\cite{bazzani2016self,simonyan2013deep,singh2017hide,zhou2016learning,zhang2018adversarial,zhang2018self,xue2019danet}. Such methods predict both object class and location by using only classification labels. However, since loss functions widely used in fully supervised settings are not directly generalizable to weakly supervised counterparts, it remains a challenging problem as to how to develop an effective supervision for object localization using only image-level information.

\begin{figure}[t!]
    \setlength{\abovecaptionskip}{0.0cm}
    \setlength{\belowcaptionskip}{-0.3cm}
	\centering
	\includegraphics[width=0.99\textwidth, height=0.64\textwidth]{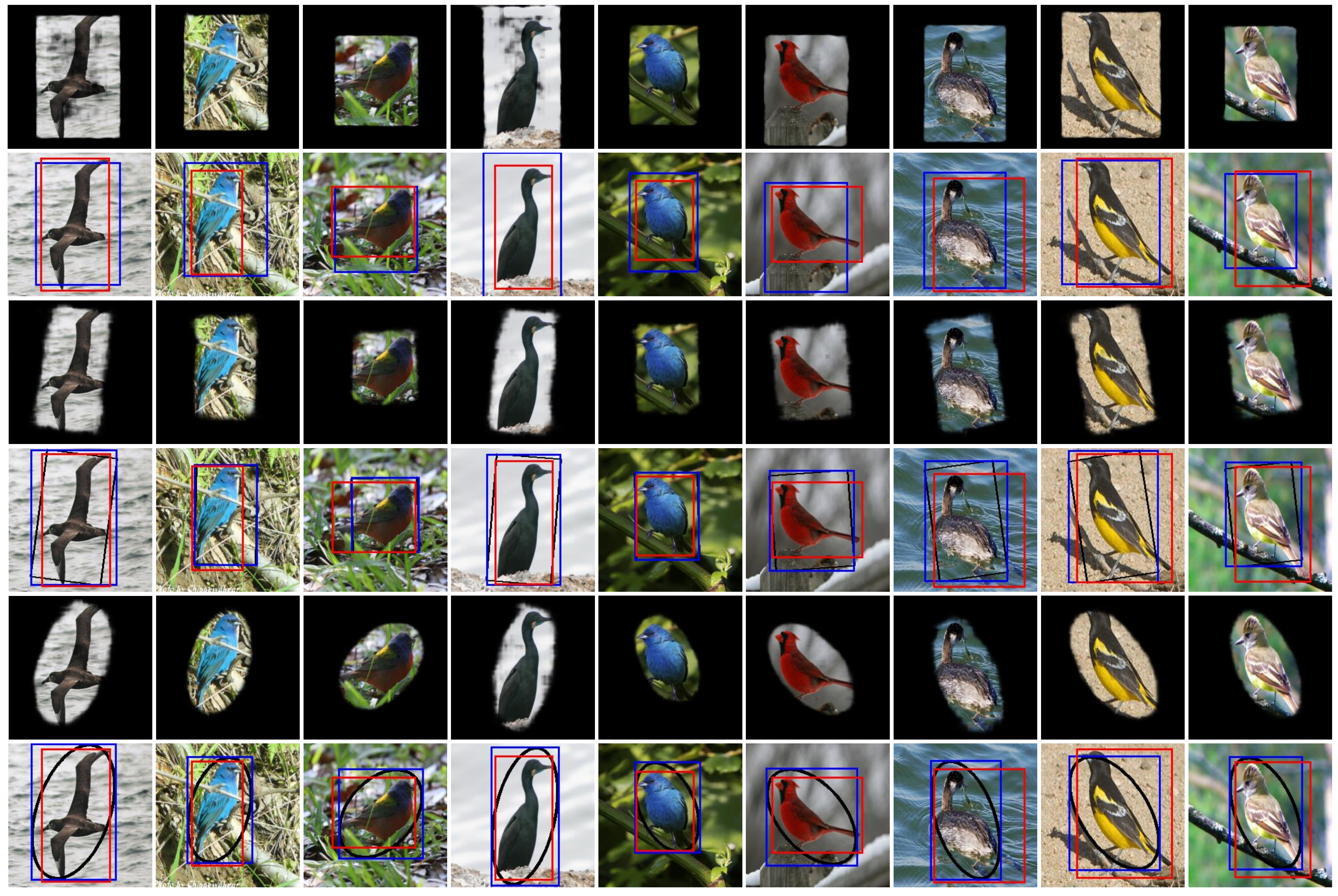}
	\caption{Weakly supervised object localization results of examples from CUB-200-2011 dataset using GC-Net. 1st-2nd rows: predictions using a normal rectangle geometry constraint. 3rd-4th rows: predictions using a rotated rectangle geometry constraint. 5th-6th rows: predictions using a rotated ellipse geometry constraint. Predicted and ground-truth bboxes are in blue and red, respectively. Rotated rectangles and ellipses are in black, which induced the predicted bboxes.}
	\label{fig:cub200}
\end{figure}

Up to update, two types of learning-based approaches are commonly used for the WSOL task, including self-taught learning \cite{bazzani2016self} and methods that take advantage of class activation maps (CAMs) \cite{zhou2016learning,zhang2018adversarial,zhang2018self,xue2019danet}. Unfortunately, the former method is not end-to-end. While the latter CAM-based approaches being able to learn end-to-end, they are suffering two obvious issues. First, the use of activated regions which sometimes are ambiguous may not be able to reflect the exact location of object of interest. As such, supervision signal produced by these methods is not strong enough to train a deep network for precise object localization. The second issue is that in these approaches a threshold value needs to be tuned manually and carefully so as to extract good bboxes from the respective activation map. 

To overcome the existing limitations above, we propose the geometry constrained network for WSOL, which we term GC-Net. It has three modules: a detector, a generator and a classifier. The detector takes responsibility for regressing the coefficients of a specific geometric shape. The generator, which can be either learning-driven or model-driven, converts these coefficients to a binary mask conforming to that shape, applied then to masking out the object of interest in the input image. This can be seen in the 1st, 3rd and 5th rows of Fig.~\ref{fig:cub200}. The classifier takes the masked images (both object and background) as inputs and performs two complementary image classification tasks. To train GC-Net effectively, we propose a novel multi-task loss function, including the area loss, the object loss and the background loss. The area loss constrains the predicted geometric shape to be tight and compact, and the object and background losses together guarantee that the masked region contains only the object. Once the network is trained using image class label information, the detector is deployed to produce object class and bbox directly and accurately. Collectively, the main contributions of the paper can be summarized as: 
\begin{itemize}
	\item We propose a novel GC-Net for WSOL in the absence of bbox annotations. Different from the currently most popular CAM-based approaches, GC-Net is trained end-to-end and does not need any post-processing step (e.g. thresholding) that may need a careful hyperparameter tuning. It is easy and accurate and therefore paves a new way to solve this challenging task.
	\item We propose a generator by learning or using knowledge about mathematical modeling. In both methods, the generator allows backpropagation of network errors. The generator also imposes a hard, explicit geometry constraint on GC-Net. In contrast to previous methods where no constraint was considered, supervision signal induced by such a geometry constraint is strong and can be used to supervise and train GC-Net effectively. 
	\item We propose three novel losses (i.e. object loss, background loss, and area loss) to supervise the training of detector. While the object loss tells the detector where the object locates, the background loss ensures the completeness of the object location. Moreover, the area loss computes the geometric shape area by imposing tightness on the resulting mask used to highlight the object for classification. These three losses work together effectively to deliver highly accurate localization.
	\item We evaluate our method on a fine-grained classification dataset, CUB-200-2011 and a large-scale dataset, ILSVRC2012. The method outperforms existing state-of-the-art WSOL methods by a large margin.
\end{itemize}

\section{Related Work}
\cite{bazzani2016self} proposed a self-taught learning-based method for WSOL, which determines the object location by masking out different regions in the image and then observes the changes of resulting classification performance. When the selected region shifts from object to background, the classification score drops significantly. This method was embedded in a agglomerative cluster to generate self-taught localization hypotheses, from which the bbox was obtained. While the changes of the classifier score indicated the location of the object, the approach was not end-to-end. Instead, localization was carried out by a follow-up clustering step.

In~\cite{zhou2016learning}, the authors proposed the global average pooling (GAP) layer to compute CAMs for object localization. Specifically, after the forward pass of a trained CNN classifier, the feature maps generated by the classifier were multiplied by the weights from the fully connected layer that uses the GAP layer outputs as the inputs. The resulting weighted feature map formed the final CAM, from which a bbox was extracted. Later on, researchers~\cite{hwang2016self,chaudhry2017discovering} found that the last fully connected layer used in the classifier in~\cite{zhou2016learning} is removable and GAP itself has the capability of classifying images. They also found that the feature maps before the GAP layer can be directly used as CAMs. As a result, these findings drastically simplified the process of generating distinctive CAMs. Although these methods are end-to-end, the use of CAMs only identifies the most distinguishing parts of the object. It is non-trivial to learn an accurate CAM that contains the compact and complete object.

Since then, different extensions have been proposed to improve the generation process of CAMs such that they can catch more complete regions belonging to the object. A self-produced guidance (SPG) approach~\cite{zhang2018self} used a classification network to learn high confident regions, under the guidance of which they then leveraged attention maps to progressively learn the SPG masks. Object regions were separated from background regions with these masks. ACoL~\cite{zhang2018adversarial} improved localization accuracy through two adversarial classification branches. In this method, different regions of the object were activated by the adversarial learning and the network inter-layer connections. The resulting CAMs from the two branches were concatenated to predict more complete regions of the object. Similarly, DA-Net~\cite{xue2019danet} used a discrepant divergent activation (DDA) to minimize the cosine similarity between CAMs obtained from different branches. Each branch can thus activate different regions of the object. For classification, DA-Net employed a hierarchical divergent activation (HDA) to predict hierarchical categories, and the prediction of a parent class enabled the activation of similar regions in its child classes.

\section{The Proposed Method}
\begin{figure}[t]
    \setlength{\abovecaptionskip}{0.0cm}
    \setlength{\belowcaptionskip}{-0.3cm}
	\centering
	\includegraphics[width=0.99\textwidth]{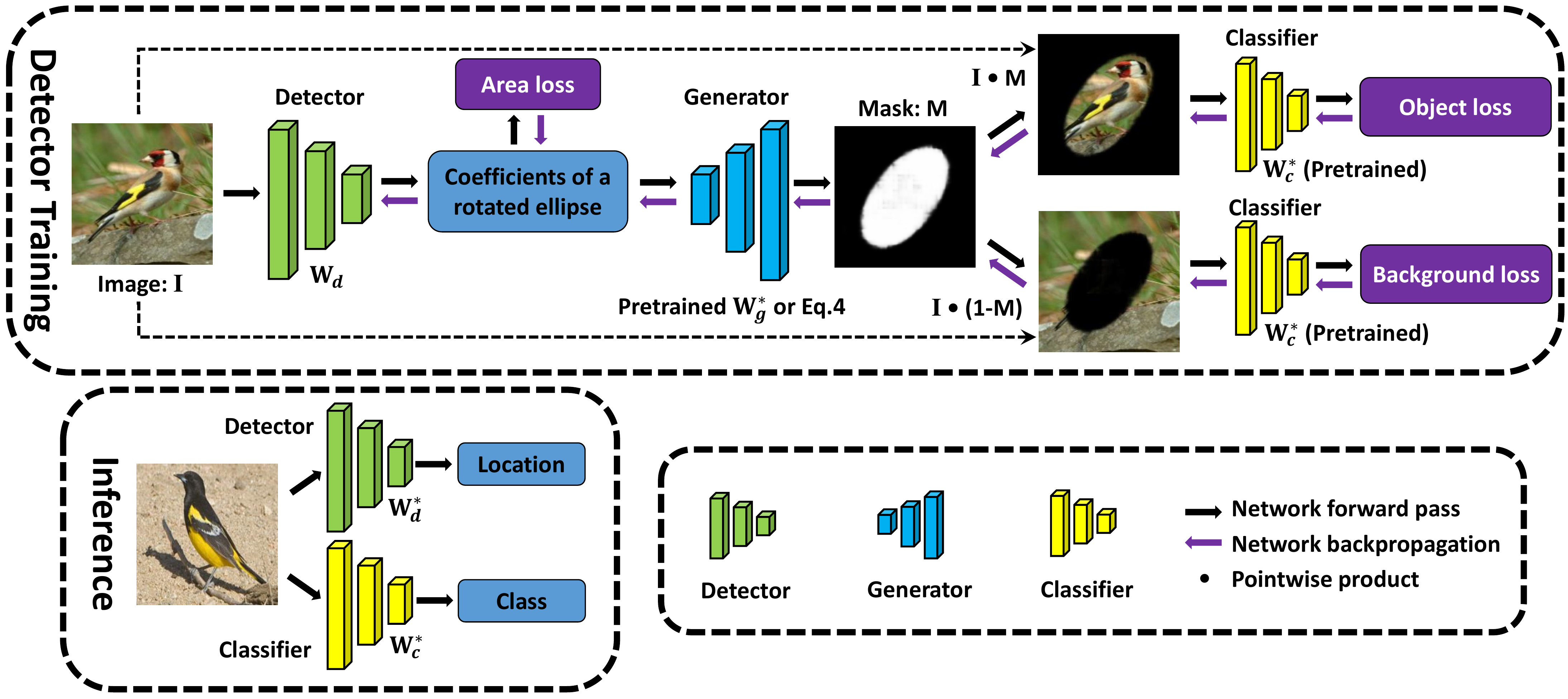}
	\caption{The architecture of our GC-Net including the detector, generator and classifier. In this figure, the geometry constraint is imposed by a rotated ellipse. The network is trained end-to-end and during inference the classifier and detector respectively predict the object category and location. No post-processing, such as thresholding, is required.}
	\label{fig:framework}
\end{figure}

Different from the methods above, Our GC-Net consists of three modules: a detector, a generator, and a classifier. The detector predicts a set of coefficients representing some geometric shape enclosing the object. The generator transforms the coefficients into a binary mask. The classifier then classifies the resulting masked images. During training, only classification labels are needed and during inference the detector is used to predict the geometry coefficients, from which object location can be computed. An overview of the proposed GC-Net is given in Fig.~\ref{fig:framework}. In the following, we provide more details about each module, as well as define the loss functions for training these modules.  

\subsection{Detector}
\label{sec:detector}
The detector can be a state-of-the-art CNN architecture for image classification, such as VGG16, GoogLeNet, etc. However, for different geometric shapes, we need to change the output number of the last fully connected layer in the detector. For example, for a normal rectangle, the detector has 4 outputs (i.e. coefficients): namely the center ($c_x$, $c_y$), the width $h$, and the height $w$. For a rotated rectangle, the detector regresses 5 coefficients: the center ($c_x$, $c_y$), the width $a$, the height $b$, and the rotation angle $\theta$. For an ellipse, the detector regresses 5 coefficients: the center ($c_x$, $c_y$), the axis $a$, the axis $b$, and the rotation angle $\theta$. In our experiments, we will compare object localization accuracy using these shape designs. Using one set of coefficients, one can easily compute a respective geometric shape, which can form a binary mask in that image. However, it is non-trivial if one wants to backpropagate network errors during training. To tackle this, next we propose two methods to generate the object mask.

\subsection{Generator}
\label{sec:genertor}
\begin{figure}[t]
    \setlength{\abovecaptionskip}{0.0cm}
    \setlength{\belowcaptionskip}{-0.3cm}
	\centering
	\includegraphics[width=0.95\textwidth]{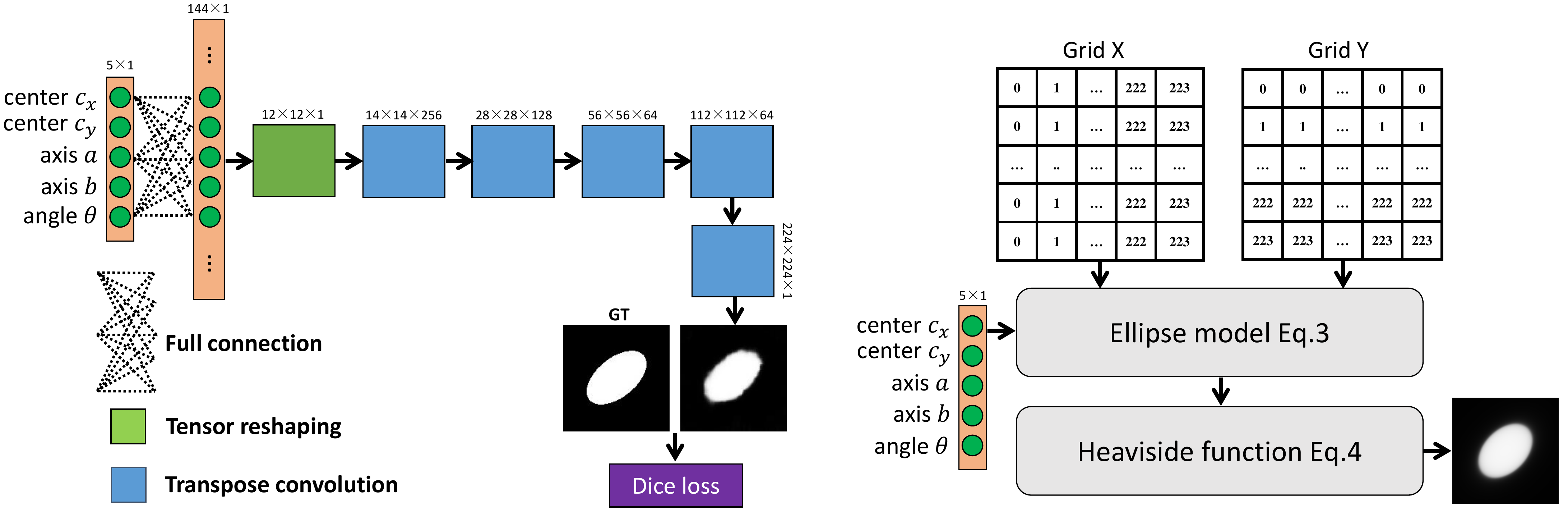}
	\caption{Object mask generation using learning-driven (left) and model-driven (right) methods. Both methods produce a mask via the coefficients regressed from the detector. }
	\label{fig:C2MC_framework}
\end{figure}

In this section, we propose a learning-driven method and a model-driven method to generate the object mask, The accuracy of each method will be compared in Section~\ref{section:self_compct}. Fig.~\ref{fig:C2MC_framework} left shows the learning-driven mask generator, which uses a network to learn the conversion between the coefficients from the detector and the object binary mask. Fig.~\ref{fig:C2MC_framework} right shows the model-driven mask generator. In this method, the conversion is done by using mathematical models (knowledge) without learning. Both methods impose a hard constraint on the detector such that the predicted shape satisfy some specific geometric constraints. Such a constraint improves localization accuracy and makes our method different from previous methods based on CAMs.

\textbf{Learning-driven generator:} The mask generator can be a neural network, such as the one defined in Fig.~\ref{fig:C2MC_framework} left, where we showed an example about how to generate a mask from the 5 coefficients of an ellipse. In the generator, the input was a 5-dimensional vector (representing the 5 coefficients), which was fed to a fully connected layer resulting in a 144-dimensional vector. The new vector was reshaped to a two-dimensional tensor (excluding the last dimension), which was then upsampled all the way to the size of the original image. The upsampling was carried out through the transpose convolution operation. The network architecture was inspired by the AUTOMAP~\cite{zhu2018image} for image reconstruction. Here, we have modified it to improve computational efficiency. 

It is necessary to pretrain the mask generator before we use it to optimize the weights in the detector. To do so, we need to generate lots of paired data for training. Using ellipse as an example, the paired data is defined as the 5 coefficients versus the ground truth binary mask corresponding to these coefficients. To generate such paired data, we randomly sampled the coefficients following a Gaussian distribution. With the Dice loss and synthesized paired data, we are able to train the mask generator. The training process and details have been given in Section \ref{sec:implementation}.

Once the generator is trained, we freeze its weights and connect it to the detector. By doing this, the generator has the capability of mapping the coefficients predicted from the detector to a binary mask, which can be employed to identify the object region as well as the background region. On the other hand, the errors produced by the classifier (introduced next) can propagate back to the detector, so the training process of the detector is effectively supervised.

\begin{figure}[t]
    \setlength{\abovecaptionskip}{0.0cm}
    \setlength{\belowcaptionskip}{-0.3cm}
	\centering
	\subfigure{\label{fig:mask_math_1}\includegraphics[width=0.4\textwidth, height=0.33\textwidth]{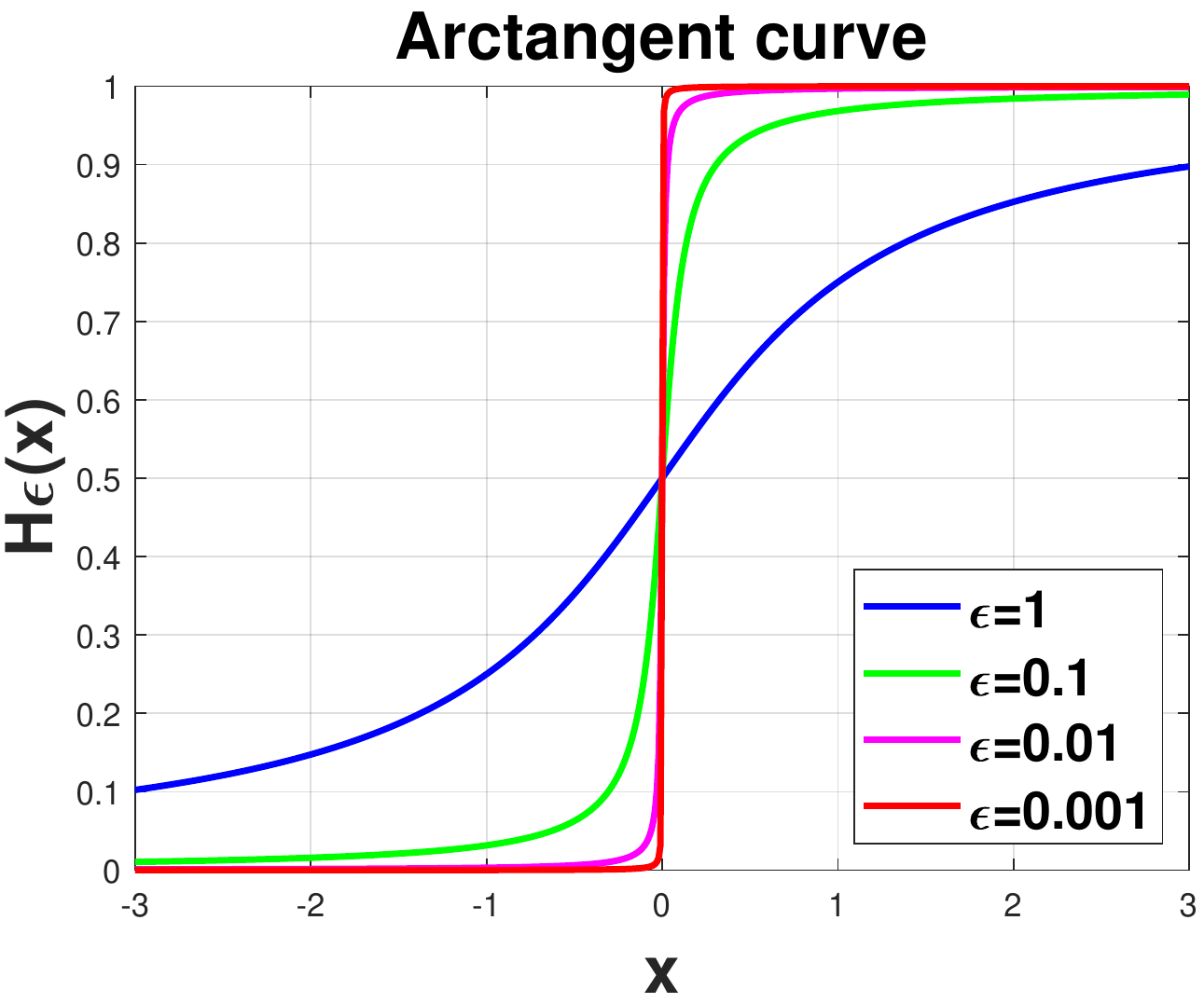}}
	\subfigure{\label{fig:mask_math_2}\includegraphics[width=0.58\textwidth]{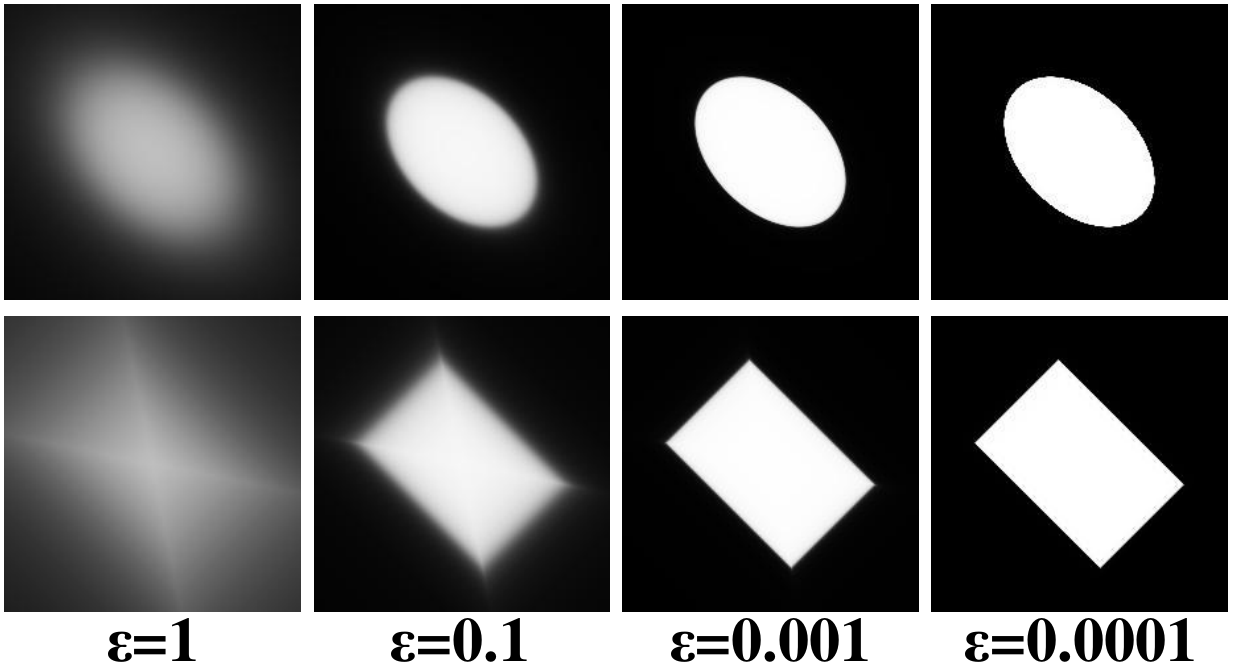}}
	\caption{Impact of $\epsilon$ on the approximated Heaviside function for 1D (left) and 2D (right) cases. The smaller $\epsilon$ is, the closer the function is approaching to the true binary function.}
	\label{fig:mask_math}
\end{figure}

\textbf{Model-driven generator:} Instead of learning, the mask generation process can be also realized by a mathematical approach. The illustration is given in Fig.~\ref{fig:C2MC_framework} right. Again let us use ellipse as an example. Similar deviations for other geometric shapes have been given in the Appendix. Given the coefficients ($c_x,c_y,\theta,a,b$) of a general ellipse, we have the following mathematical model to represent it 
\begin{equation}
    \frac{((x-c_{x})cos\theta + (y-c_{y})sin\theta)^2}{a^2} + \frac{((x-c_{x})sin\theta - (y-c_{y})cos\theta)^2}{b^2} = 1,
\end{equation}
where $x,y:\rm{\Omega} \subset \mathbb{R}^2$. To generate the mask induced by the ellipse, we can use the following Heaviside (binary) function
\begin{equation} \label{eq:H}
    H(\phi(x,y)) = \left\{\begin{array}{ll}
      0 &  \;\; \phi(x,y) >0 \\
      1 &  \;\; \phi(x,y) \le 0\\
\end{array} \right.,
\end{equation}
with $\phi(x,y)$ defined as 
\begin{equation} 
    \phi(x,y) = \frac{((x-c_{x})cos\theta + (y-c_{y})sin\theta)^2}{a^2} + \frac{((x-c_{x})sin\theta - (y-c_{y})cos\theta)^2}{b^2} - 1.
\end{equation}
However, it is difficult to backpropagate network errors using such a representation due to its non-differentiability. To tackle this difficulty, we propose to use the inverse of a tangent function to approximate the Heaviside function \cite{chan2001active,duan2014some}
\begin{equation} \label{eq:tangent}
   H_{\epsilon}(\phi(x,y)) = \frac{1}{2}\left(1+\frac{2}{\pi}arctan\left(\frac{\phi(x,y)}{\epsilon}\right)\right). 
\end{equation}
With this equation, the pixel indices falling in the ellipse region are 1, otherwise 0. Note that there is a hyperparameter $\epsilon$ which controls the smoothness of the Heaviside function. The bigger its value is, the smoother $H_{\epsilon}$ will be. When $\epsilon$ is infinitely close to zero, \eqref{eq:tangent} is equivalent to \eqref{eq:H}. In Fig.~\ref{fig:mask_math}, we illustrate the results of using different $\epsilon$ in this approximation for 1D and 2D cases. In our implementation, we made the parameter learnable in our network in order to avoid manual tuning of this hyperparameter. Of note, if the generator is chosen model-driven, we can use it directly in GC-Net in Fig.~\ref{fig:framework} without pretraining. 

\subsection{Classifier}
The classifier is a common image classification neural network. In inference phase, the classifier is responsible for predicting the image category. In detector training phase, it takes the resulting masked images as inputs and performs two complementary classification tasks: one for object region and another for background region. Similarly to the learning-driven generator, we need to pretrain the classifier before we use it to optimize the weights in the detector. The classifier could be pretrained using ILSVRC2012~\cite{ILSVRC15} and then fine tuned to recognize the objects in the detection context. The loss function used was the categorical cross-entropy loss, defined in loss (\ref{eq:objectloss}).  Once the classifier is trained,  we freeze its weights and connect it to the generator. 

\subsection{Loss functions}
\label{sec:loss}
After both the object mask generator and classifier are pretrained (if the generator mode is learning-driven), we can start to optimize the weights in the detector. Three loss functions were developed to supervise the detector training: the area loss ${\cal L}_a$, the object loss ${\cal L}_o$, and the background loss ${\cal L}_b$. In  Fig.~\ref{fig:framework}, we have illustrated where they should be used. The final loss is a sum of the three, given as
\begin{equation}
\label{eq:loss}
{\cal L}(\textbf{W}_d) = \alpha{\cal L}_a(\textbf{W}_d) + \beta{\cal L}_{o}(\textbf{W}_d, \textbf{W}_g^*, \textbf{W}_c^*) + \gamma{\cal L}_{b}(\textbf{W}_d, \textbf{W}_g^*, \textbf{W}_c^* ),  
\end{equation}
where $\alpha$, $\beta$ and $\gamma$ are three hyperparameters balancing the three losses. $\textbf{W}_d$ denotes the network weights in the detector; $\textbf{W}_g^*$ denotes the fixed, pretrained network weights in the mask generator; $\textbf{W}_c^*$ denotes the fixed, pretrained network weights in the classifier. The aim is to find the optimal $\textbf{W}_d^*$ such that the combined loss is minimized.
Here the area loss is imposed on the object mask. This loss ensures the tightness/compactness of the mask, without which the mask size is not constrained and therefore can be very big sometimes. The area of a geometrical shape can be simply approximated by
\begin{equation}
{\cal L}_{a} = a \cdot b,
\label{area_loss}
\end{equation}
where $\cdot$ denotes the pointwise product; $a$ and $b$ can represent the two axes of an ellipse or the width and height of a rectangle, which are two output coefficients from the detector. 

Next, the object loss is defined as the following categorical cross-entropy
\begin{equation} \label{eq:objectloss}
{\cal L}_{o} = -\sum_j^m \sum_i^n q_{i,j} {\rm{log}} \left(\frac{{e}^{p^o_{i,j}}}{\sum_k^n{{e}^{p^o_{k,j}}}} \right),\\
\end{equation}
where $m$ and $n$ denote the number of training samples and the number of class labels, respectively; $q$ stands for the ground truth class label; ${p^o}$ represents the output of the classifier fed with the masked image enclosing only object region, and it is of the form 
\begin{equation} \nonumber
{p^o} = \textbf{CNN}(M \cdot I, \{\textbf{W}_d, \textbf{W}_g^*, \textbf{W}_c^*\}).     
\end{equation}
$\textbf{CNN}$ above represents the whole network with the weights $\{\textbf{W}_d, \textbf{W}_g^*, \textbf{W}_c^*\}$ and it takes as input the original image $I$ multiplied by the mask $M$. 

The value of the object loss (\ref{eq:objectloss}) is small if the object is enclosed correctly inside the mask $M$. However, using this loss alone, we found in experiments that the masked region sometimes contains the object partially, such as head or body of a bird. This observation motives us to consider how to use background region. As such, we propose the following background loss
\begin{equation} \label{eq:backgroundloss}
{\cal L}_{b} =  \sum_j^m \sum_i^n \frac{{e}^{p^b_{i,j}}}{\sum_k^n{{e}^{p^b_{k,j}}}}  {\rm{log}} \left( \frac{{e}^{p^b_{i,j}}}{\sum_k^n{{e}^{p^b_{k,j}}}}  \right), \\
\end{equation}
where ${p^b}$ represents the output of the classifier fed with the masked image enclosing only background region, and it is of the form
\begin{equation} \nonumber
{p^b} = \textbf{CNN}( I \cdot (1-M), \{\textbf{W}_d, \textbf{W}_g^*, \textbf{W}_c^*\}).  
\end{equation}

\begin{figure}[t]
    \setlength{\abovecaptionskip}{0.0cm}
    \setlength{\belowcaptionskip}{-0.3cm}
	\centering
	\subfigure{\label{fig:entropy_1}\includegraphics[scale=0.27]{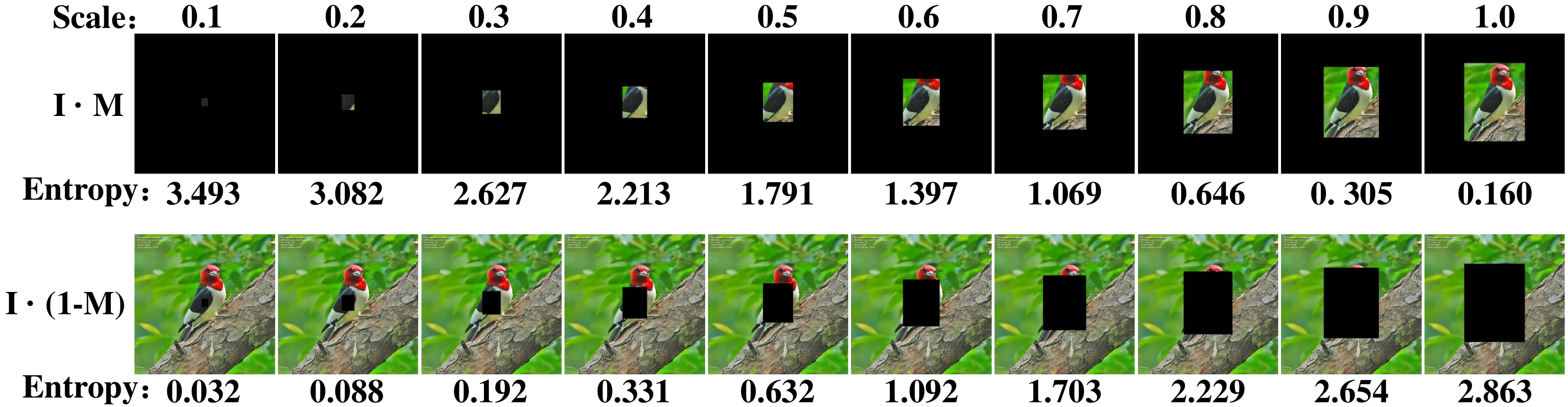}}
	\subfigure{\label{fig:entropy_2}\includegraphics[scale=0.22]{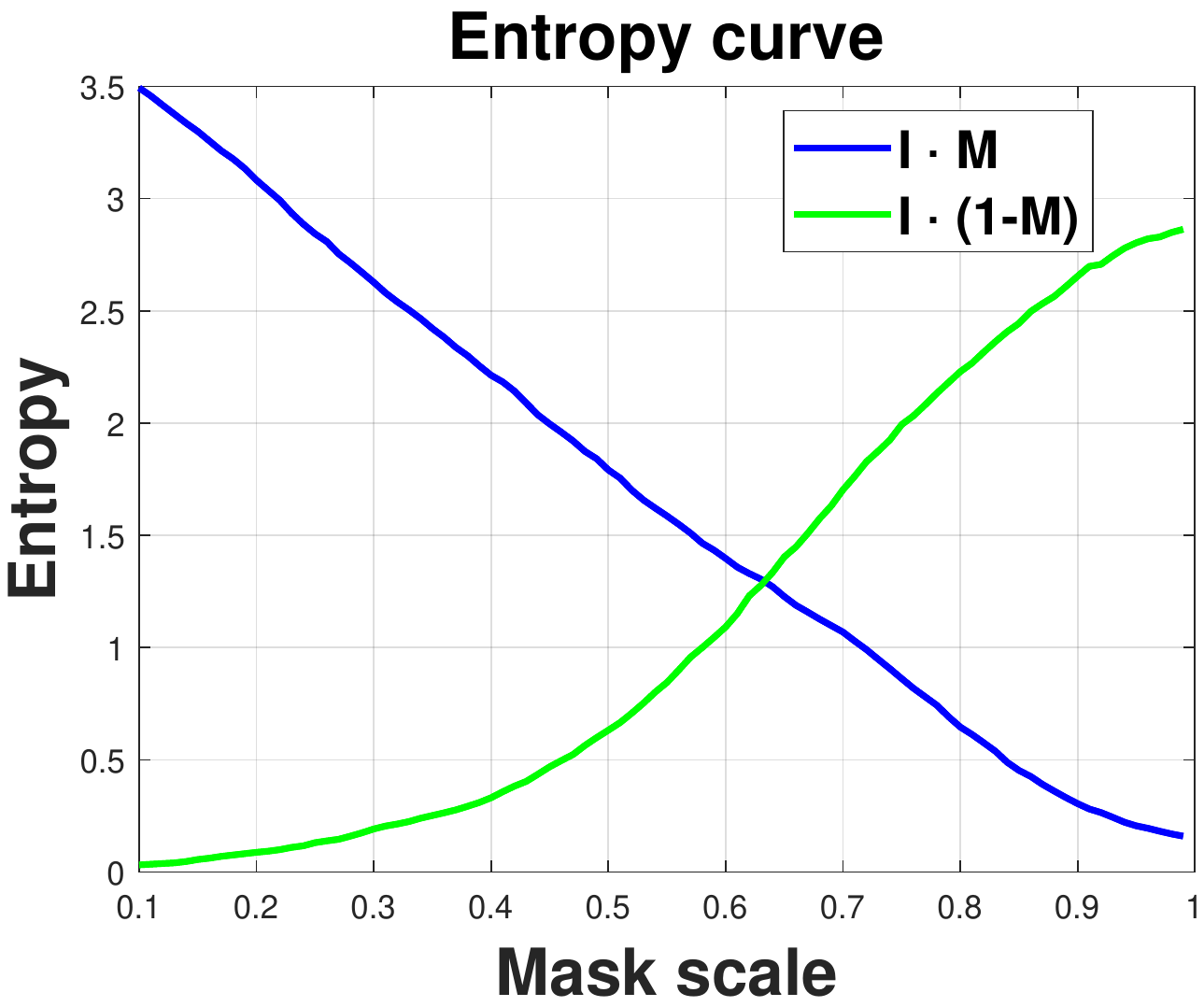}}
	\caption{Statistical analysis of entropy. Left: masked object regions (top) and background regions (bottom). Right: entropy values versus mask scales. The entropy increases when there are more uncertainties. Oppositely, it decreases when more certainties are present.}
	\label{fig:entropy}
\end{figure}

We note that the proposed loss (\ref{eq:backgroundloss}) is known as the negative entropy. In information theory, entropy is the measure of uncertainty in a system or an event. In our case, we want the classifier to produce the maximum uncertainty on the background region, which is the situation that only pure background remains (i.e. the object is completely enclosed by the mask, as shown in Fig.~\ref{fig:entropy}). The negative sign is to reverse the maximum entropy to the minimum. By minimizing the three loss functions simultaneously, we are able to classify the object accurately and meanwhile produce compact and complete bbox around the object of interest using only classification labels. Our method is end-to-end without using any post-processing step and therefore is very accurate, as can be confirmed from our experiments next. 

\section{Experiments}
In this section, we first introduce datasets and quantitative metrics used for experiments. This is followed by implementation details of the proposed method as well as ablation studies of different loss functions. Next, learning- and model-driven methods are compared and different geometry constraints are evaluated. Extensive comparisons with state-of-the-art methods are given in the end. 

\subsection{Datasets and evaluation metrics}
We evaluated our GC-Net using two large-scale datasets, i.e., CUB-200-2011~\cite{wah2011caltech} and ILSVRC2012~\cite{ILSVRC15}. CUB-200-2011 is a fine-grained classification dataset with 200 categories of birds. There are a total of 11,788 images, which were split into 5,994 images for training and 5,794 images for testing. For the ILSVRC2012 dataset, we chose the subset\footnote{This subset has not been changed or modified since 2012.} where we have ground truth labels for this WSOL task. The subset contains 1000 object categories, which have already been split into training and validation. We used 1.2 million images in the training set to train our model and 50,000 images in the validation set for testing.

For evaluation metrics, we follow~\cite{deselaers2012weakly} and~\cite{ILSVRC15}, where they defined the location error~\cite{deselaers2012weakly} and the correct localization~\cite{ILSVRC15} for performance evaluation. The location error (LocErr) is calculated based on both classification and localization accuracy. More specifically, LocErr is 0 if both classification and localization are correct, otherwise 1. Classification is correct if the predicted category is the same to ground truth, and localization is correct if the value of intersection over union (IoU) between the predicted bbox and the ground truth bbox is greater than 0.5. The smaller the LocErr is, the better the network performs. The correct location (CorLoc) is computed solely based on localization accuracy. For example, it is 1 if IoU$>0.5$. The higher the CorLoc is, the better the method is. In some experiments, we also reported the classification error (ClaErr) for performance evaluation.

\subsection{Implementation details}
\label{sec:implementation}
We need to pretrain the generator and classifier prior to training the detector. We first provide implementation details of training the classifier. For ILSVRC2012, we directly used the pretrained weights provided by PyTorch for the classifier. For CUB-200-2011, we changed the output size of the classifier from 1000 to 200 and initialized remaining weights using those pretrained from ILSVRC2012. We then fine tuned the weights on CUB-200-2011 using SGD~\cite{sutskever2013importance} with a learning rate of 0.001 and a batch size of 32.

For the learning-driven generator, we used SGD with a learning rate of 0.1 and a batch size of 128. We used Dice as the loss function as it is able to ease the class imbalance problem in segmentation. We randomly sampled many sets of coefficients, each being a $5 \times 1$ vector and representing the center ($c_x$, $c_y$), the axis $a$, the axis $b$ and the rotation angle $\theta$ (ranging from -90$^{\circ}$ to 90$^{\circ})$. These vectors and their respective masks were then fed to the generator for training. We optimized the generator for 0.12 million iterations and within each iteration we used a batch size of 128. By the end of training the generator has seen 15 million paired data and therefore is generalizable enough to unseen coefficients.     

To train the detector, we freezed the weights of the pretrained generator and classifier and updated the weights only in the detector. We used Adam~\cite{kingma2014adam} optimizer with a learning rate of 0.0001, as we found that it is difficult for SGD to optimize the detector effectively. For CUB-200-2011, we used a batch size of 32. For ILSVRC2012, we used a batch size of 256. The detector outputs were activated by the sigmoid nonlinearity before they were passed to the generator. We tested several commonly used backbone network architectures, including VGG16~\cite{simonyan2014very}, GoogLeNet~\cite{szegedy2015going} and Inception-V3~\cite{szegedy2016rethinking}. Of note, we used the same backbone for both the classifier and detector.

\subsection{Ablation studies}
\begin{figure}[h!]
    \setlength{\abovecaptionskip}{0.0cm}
    \setlength{\belowcaptionskip}{-0.3cm}
	\centering
	\subfigure{\label{fig:ablation}\includegraphics[scale=0.32]{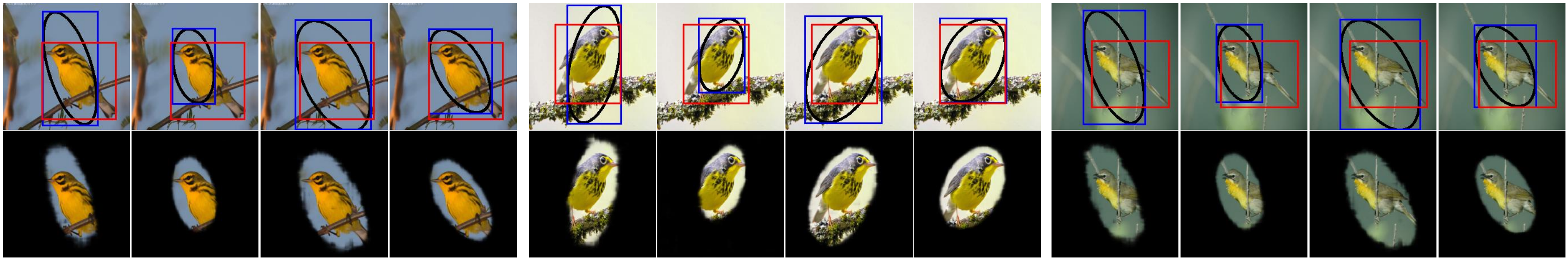}}
	\caption{Three examples showing the impact of using different losses. The predicted (blue) and ground-truth (red) bboxes are shown in top row. For each example, from left to right the losses used are ${\cal {L}}_o$, ${\cal {L}}_o$+${\cal {L}}_a$, ${\cal {L}}_o$+${\cal {L}}_b$ and ${\cal {L}}_a$+${\cal {L}}_o$+${\cal {L}}_b$, respectively.}
	\label{fig:loss}
\end{figure}

The ablation studies on CUB-200-2011 were performed to evaluate the contribution of each loss (i.e. the area loss ${\cal {L}}_a$, the object loss ${\cal {L}}_o$ and the background loss ${\cal {L}}_b$ in Section~\ref{sec:loss}) for localization. For this experiment, we trained CG-Net using VGG16 as backbone and the learning-driven generator constrained by the rotated ellipse. Table~\ref{tab:loss_compare} reported the localization accuracy measured by LocErr and CorLoc. When only ${\cal {L}}_o$ was used, there are two obvious issues: (1) CG-Net could not guarantee the mask is tight and compact to get rid of irrelevant background regions; and (2) CG-Net fails to detect some regions belonging to the object, These issues can be clearly observed in the 1st column of each example in Fig.~\ref{fig:loss}.

\begin{wraptable}{r}{0.4\textwidth}
    \setlength{\abovecaptionskip}{-0.7cm}
    \setlength{\belowcaptionskip}{0.1cm}
	\centering
	\caption{Comparison of the object localization performance on CUB-200-2011 using different losses.}
    \resizebox{0.4\textwidth}{!}{
	\begin{tabular}{lrrr}  
		\toprule
		&\multicolumn{2}{c}{LocErr}\\
		\cmidrule(lr){2-3}
		Loss functions & Top1 & Top5 & CorLoc \\
		\midrule
		${\cal {L}}_o$            & 59.22 & 51.75 & 51.69 \\
	${\cal {L}}_o$+${\cal {L}}_a$       & 69.89 & 63.12 & 39.89 \\
	${\cal {L}}_o$+${\cal {L}}_b$       & 47.03 & 37.69 & 66.52 \\
	${\cal {L}}_a$+${\cal {L}}_o$+${\cal {L}}_b$   & \textbf{41.15} & \textbf{30.10} & \textbf{74.89} \\
		\bottomrule
	\end{tabular}}
	\label{tab:loss_compare}
	\vspace{-10pt}
\end{wraptable}
To address the first issue, we added ${\cal {L}}_a$ to penalize area such that irrelevant background can be removed.  However, using ${\cal {L}}_o$+${\cal {L}}_a$ made the network focus on the most discriminate regions, as shown in the 2nd column of each example in Fig.~\ref{fig:loss}. This side effect led to a sharp decreasing in localization accuracy, suggested by both LocErr and CorLoc (39.89\%) in Table~\ref{tab:loss_compare}. This is because in many cases GC-Net detected only very small regions such as upper bodies of birds, reducing the IoU value and hence resulting in a big accuracy drop. As such, it is necessary to address the second issue. For this, we further added ${\cal {L}}_b$. This loss maximizes the uncertainty for background classification and therefore compensates the incomplete localization problem. As shown in the last column of each example in Fig.~\ref{fig:loss} and the last row in Table~\ref{tab:loss_compare}, such a combination delivered the most accurate performance. From the figure, we can also see that the localization contains more irrelevant background if only ${\cal {L}}_o + {\cal {L}}_b$ is used without the area constraint. This experiment proved that all three loss functions are useful and necessary.

\subsection{Learning-driven versus model-driven geometry constraints}
\label{section:self_compct}

In this section, we want to test which generator is better: model-driven or learning-driven? Also, we intend to see the performance of using different geometry constraints. As such, we performed experiments on CUB-200-211 using VGG16 as backbone for the detector. For each geometry, we implemented both learning- and model-driven strategies. 

Table~\ref{tab:Mask_compar} reported location errors using both generators under different geometry constraints. The LocErr from the learning-driven generator was higher than that from the model-driven generator. During experiments, we found that the model-driven approach was more sensitive to hyperparameter tuning (i.e. $\alpha$, $\beta$ and $\gamma$ in the loss). In addition, different initializations of learnable $\epsilon$ in Eq.~(\ref{eq:tangent}) also affected localization accuracy a lot. Through many attempts, $\epsilon$ was initialized to 0.1, and $\alpha$, $\beta$ and $\gamma$ were set to 1, 2.5 and 1 respectively. In contrast, the learning-driven method was robust to hyperparameter tuning. We were able to get a decent performance by setting both $\alpha$, $\beta$ and $\gamma$ to 1. As such, we think that the inferior performance of the model-driven method may be due to the difficulty of hyperparameter tuning. Its performance may be further boosted by a more careful hyperparamter search. Although the learning-driven method has the advantage of a better localization performance, the model-driven method does not need training in advance.
\begin{table}[t]
	\setlength{\abovecaptionskip}{0.1cm}
    \setlength{\belowcaptionskip}{-0.3cm}
	\centering
	\renewcommand\tabcolsep{10pt}
	\resizebox{0.99\textwidth}{!}{
	\begin{tabular}{llrrrc}  
		\toprule
		& &\multicolumn{2}{c}{LocErr}\\
		\cmidrule(lr){3-4}
Strategies	&	Geometries  & Top1 & Top5 & CorLoc & Rotation\\
		\midrule
	&	Rectangle           & 44.35 & 34.25 & 70.06 & $\times$\\
	Learning-driven &	Rotated rectangle  & \textbf{36.76} & \textbf{24.46} & \textbf{81.05} & \checkmark \\
	&	Rotated ellipse    & 41.15 & 30.10 & 74.89 & \checkmark   \\
		\midrule
	&	Rectangle          & 44.10 & 33.28 & 71.56 & $\times$\\
	Model-driven &	Rotated rectangle  & 44.25 &33.62 & 71.13 & \checkmark \\
	&	Rotated ellipse    & 41.73 & 30.60 & 74.61 & \checkmark   \\
		\bottomrule
	\end{tabular}}
	\caption{Comparison of the object localization performance on CUB-200-2011, using learning-driven and model-driven generators under different geometry constraints.}
	\label{tab:Mask_compar}

\end{table}

Table~\ref{tab:Mask_compar} also reported location errors using three masks with different geometrical shapes. The least accurate geometry was rectangle, because it can easily include irrelevant background regions, thus decreasing the overall performance of the detector. Moreover, normal rectangles were unable to capture rotations, which seemed to be crucial to compute a high IOU value. In contrast, rotated rectangles were able to filter out noisy background regions and achieved the best localization performance among all three geometries. Although the localization accuracy from rotated ellipses was between rectangles and its rotated versions, they achieved the best performance in predicting rotations, which can be confirmed in the last two rows of Fig.~\ref{fig:cub200}. As is evident, rotations predicted by ellipses have a better match with true rotations of the objects. In contrast, rotations predicted by rotated rectangles were less accurate, as shown in the 3th and 4th rows of Fig.~\ref{fig:cub200}. Due to the lack of ground truth rotation labels, we could not study rotation quantitatively.

\subsection{Comparison with the state-of-the-art}
In this section, we compared our GC-Net and its variants with the existing state-of-the-art on CUB-200-2011 and ILSVRC2012. We used VGG16, GoogLeNet and Inception-V3 as three backbones. Table~\ref{tab:cub200_compar} and~\ref{tab:ImageNet_compar} reported the performance of different methods.
\begin{table}[t]
    \setlength{\abovecaptionskip}{0.1cm}
    \setlength{\belowcaptionskip}{0cm}
	\centering
	\renewcommand\tabcolsep{6pt}
	\resizebox{0.7\textwidth}{!}{
	\begin{tabular}{lrrrrr}  
		\toprule
		&\multicolumn{2}{c}{ClsErr} &\multicolumn{2}{c}{LocErr}\\
		\cmidrule(lr){2-3} \cmidrule(lr){4-5}
		Methods compared & Top1 & Top5 & Top1 & Top5 & CorLoc\\
		\midrule
		CAM-VGG~\cite{zhou2016learning}                   & 23.4 & \textbf{7.5} & 55.85 & 47.84 & 56.0    \\
		ACoL-VGG~\cite{zhang2018adversarial}                  & 28.1 &  -  & 54.08 & 43.49 &54.1     \\
		SPG-VGG~\cite{zhang2018self}                   & 24.5 & 7.9 & 51.07 & 42.15 & 58.9 \\
		TSC-VGG~\cite{he2017weakly}   & - & - & -& - & 65.5      \\
		DA-Net-VGG~\cite{xue2019danet}                 & 24.6 & 7.7 & 47.48 & 38.04 &67.7     \\
		GC-Net-Elli-VGG (ours)          & \textbf{23.2} & 7.7 & \textbf{\textcolor[rgb]{0.5,0.5,0.5}{41.15}} & \textbf{\textcolor[rgb]{0.5,0.5,0.5}{30.10}} & \textbf{\textcolor[rgb]{0.5,0.5,0.5}{74.9}}    \\
		GC-Net-Rect-VGG (ours)          & \textbf{23.2} & 7.7 & \textbf{36.76} & \textbf{24.46} &\textbf{81.1}    \\
		\cmidrule(lr){1-6}
		CAM-GoogLeNet~\cite{zhou2016learning}            & 26.2 & 8.5 & 58.94 & 49.34& 55.1     \\
		Friend or Foe-GoogLeNet~\cite{xu2016friend}       &-&-&-&-& 56.5      \\
		SPG-GoogLeNet~\cite{zhang2018self}            &   -  &  -  & 53.36 & 42.28 & -    \\
		DA-Net-Inception-V3~\cite{xue2019danet}           & 28.8 & 9.4 & 50.55 & 39.54& 67.0    \\
		GC-Net-Elli-GoogLeNet (ours)    & \textbf{23.2} & \textbf{6.6} & \textbf{\textcolor[rgb]{0.5,0.5,0.5}{43.46}} & \textbf{\textcolor[rgb]{0.5,0.5,0.5}{31.58}} & \textbf{\textcolor[rgb]{0.5,0.5,0.5}{72.6}}  \\
		GC-Net-Rect-GoogLeNet (ours)    & \textbf{23.2} & \textbf{6.6} & \textbf{41.42} & \textbf{29.00}& \textbf{75.3}      \\
		\bottomrule
	\end{tabular}}
	\caption{Comparison of the performance between GC-Net and the state-of-the-art on the CUB-200-2011 test set. Our method outperforms all other methods by a large margin for object localization. Here `ClsErr', `LocErr' and 'CorLoc' are short for classification error, location error and correct location, respectively.}
	\label{tab:cub200_compar}
\end{table}

\begin{table}[h]
    \setlength{\abovecaptionskip}{0.1cm}
    \setlength{\belowcaptionskip}{-0.3cm}
	\centering
	\renewcommand\tabcolsep{10pt}
	\resizebox{0.7\textwidth}{!}{
	\begin{tabular}{lrrrr}  
		\toprule
		&\multicolumn{2}{c}{ClsErr} &\multicolumn{2}{c}{LocErr}\\
		\cmidrule(lr){2-3} \cmidrule(lr){4-5}
		Methods compared  & Top1 & Top5 & Top1 & Top5 \\
		\midrule
		Backprop-VGG~\cite{simonyan2013deep}      & - & - & 61.12 & 51.46     \\
		CAM-VGG~\cite{zhou2016learning}           & 33.4 & 12.2 & 57.20 & 45.14     \\
		ACol-VGG~\cite{zhang2018adversarial}          & 32.5 & 12.0 & 54.17 & 40.57     \\
		\cmidrule(lr){1-5}
		Backprop-GoogLeNet~\cite{simonyan2013deep}        & - & - & 61.31 & 50.55     \\
		GMP-GoogLeNet~\cite{zhou2016learning}             & 35.6 & 13.9 & 57.78 & 45.26     \\
		CAM-GoogLeNet~\cite{zhou2016learning}              & 35.0 & 13.2 & 56.40 & 43.00     \\
		HaS-32-GoogLeNet~\cite{singh2017hide}          & - & - & 54.53 & -     \\
		ACol-GoogLeNet~\cite{zhang2018adversarial}            & 29.0 & 11.8 & 53.28 & 42.58     \\
		SPG-GoogLeNet~\cite{zhang2018self}             & - & - & 51.40 & 40.00     \\
		DA-Net-InceptionV3~\cite{xue2019danet}           & 27.5 & 8.6 & 52.47 & 41.72     \\
		GC-Net-Elli-Inception-V3 (ours)  & \textbf{22.6} & \textbf{6.4} & 51.47 & 42.58     \\
		GC-Net-Rect-Inception-V3 (ours)  & \textbf{22.6} & \textbf{6.4} & \textbf{50.94} & 41.91     \\			
		\bottomrule
	\end{tabular}}
	\caption{Comparison of the performance between GC-Net and the state-of-the-art on the ILSVRC2012 validation set. Our methods again perform the best.}
	\label{tab:ImageNet_compar}
\end{table}
Table~\ref{tab:cub200_compar} reported the performance of our GC-Net and other methods on CUB-200-2011. We used an average result from 10 crops to compute ClsErr and the center crop to compute LocErr, which is in line with what DA-Net~\cite{xue2019danet} has done. When VGG16 was used as backbone, GC-Net constrained by the rotated rectangle (GC-Net-Rect-VGG) was the most accurate method among all compared. For top 1 LocErr, GC-Net-Rect-VGG was about 11\% lower than DA-Net-VGG. For top 5 LocErr, it was about 14\% lower than DA-Net-VGG. When GoogLeNet was used as backbone, GC-Net-rect-GoogLeNet achieved 41.42\% top 1 LocErr and 29.00\% top 5 LocErr, outperforming DA-Net-Inception-V3 by 9\% and 11\%, respectively. In terms of ClsErr, GC-Nets achieved comparable performance with CAM-based methods when VGG16 was concerned. However, GC-Net became significantly better when GoogLeNet was used. 
\begin{figure}[t]
    \setlength{\abovecaptionskip}{-0.3cm}
    \setlength{\belowcaptionskip}{-0.3cm}
	\centering
	\includegraphics[scale=0.32]{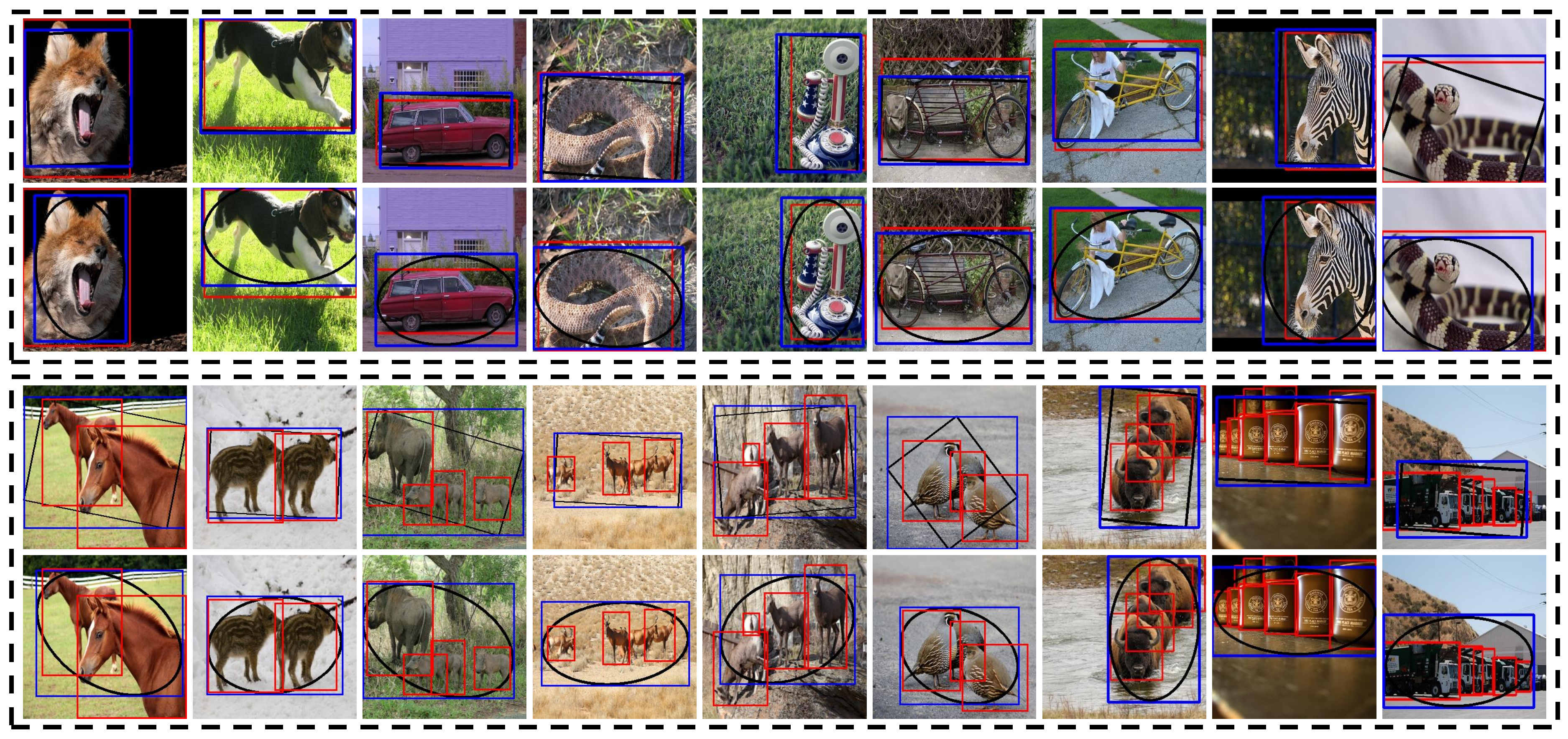}
	\caption{Localization results on some images from the ILSRC2012 dataset using GC-Net. Top: single object localization. Bottom: multiple object localization. GC-Net tends to predict a bbox that contains all target objects. Ground truth bboxes are in red, predictions are in blue. Rotated rectangles and ellipses are in black, which induced the predicted bboxes.
	}
	\label{fig:imageNet}
\end{figure}

As LocErr was calculated based on both classification and localization accuracy, a wrong classification could turn a correct localization to a wrong one. As such, in order to exclude the effect of classification, we also computed CorLoc, which is determined by only localization accuracy. In Table~\ref{tab:cub200_compar}, we reported the performance of different methods using the CorLoc metric. One can clearly see that our approach has a significantly higher CorLoc value than that of runner-up DA-Nets. The accuracy (81.1\%) of our GC-Net-Rect-VGG was about 13\% higher than that of DA-Net-VGG, and the accuracy (75.3\%) of GC-Net-rect-GoogLeNet was about 8\% higher than that of DA-Net-Inception-V3. 

For ILSVRC2012, we used inception-V3 as our backbone, which is the same for DA-Net. In order to directly use the pretrained model for our classifier, the input size of each image was resized to 299$\times$299. Table~\ref{tab:ImageNet_compar} reported the performance of GC-Nets. First, our approach obtained a much lower ClsErr than that of DA-Net, i.e., about 5\% improvement in top 1 accuracy has been achieved. However, the LocErr values of our GC-Nets were close to those of DA-Nets. Notice that on CUB-200-2011 each image contains only a single object, a large number of images in ILSVRC2012 contain multiple objects, as shown in Fig.~\ref{fig:imageNet} bottom. Overall, our methods achieved much better performance than CAM-based approaches.

\section{Conclusion}
In this study, we proposed a geometry constrained network for the challenging task of weakly supervised object localization. We have provided technical details about the proposed method and extensive numerical experiments have been carried out to evaluate and prove the effectiveness of the method. We believe that our new method will open a new door for researches in this area. 

\section*{Acknowledgement} 
The work is supported by the National Natural Science Foundation of China under Grant 91959108, 61672357 and 61602315, and  by the Ramsay Research Fund from the School of Computer Science at the University of Birmingham. 

\section{Appendix}

\subsection{Mathematical Models of Different Geometric Shapes}
In section \ref{sec:genertor}, we have defined the mathematical model of ellipse for the model-driven mask generator. We introduce here the mathematical models for three geometric shapes (i.e. rectangle, rotated rectangle, and rotated ellipse) in detail.

\noindent\textbf{Rotated ellipse}:
Given the coefficients ($c_x, c_y, \theta, a, b$) of a rotated ellipse, the mathematical model of a rotated ellipse can be defined as
\begin{equation}
\phi(x,y) = \frac{((x-c_{x})cos\theta + (y-c_{y})sin\theta)^2}{a^2} + \frac{((x-c_{x})sin\theta - (y-c_{y})cos\theta)^2}{b^2} - 1,
\end{equation}
where $x, y:\rm{\Omega} \subset \mathbb{R}^2$.

\noindent\textbf{Rectangle}: Given the coefficients ($c_x, c_y, a, b$) of a rectangle, we can represent a rectangle with the mathematical model defined as below 
\begin{equation}
\label{math:rectangle}
\phi(x,y) = \left\vert\frac{x-c_{x}}{a} + \frac{y-c_{y}}{b}\right\vert +\left\vert\frac{x-c_{x}}{a} - \frac{y-c_{y}}{b}\right\vert - 1.  
\end{equation}

\noindent\textbf{Rotated rectangle}: Given the coefficients ($c_x, c_y, \theta, a, b$) of a  rotated rectangle, the mathematical model of a rotated rectangle can be defined as 

\begin{equation}
\label{math:rotated_rectangle}
\begin{split}
\phi(x,y) =  &\left\vert\frac{(x-c_{x})cos\theta - (y-c_{y})sin\theta}{a} + \frac{(x - c_{x})sin\theta + (y-c_{y})cos\theta}{b}\right\vert +\\&\left\vert\frac{(x-c_{x})cos\theta - (y-c_{y})sin\theta}{a} - \frac{(x - c_{x})sin\theta + (y-c_{y})cos\theta}{b}\right\vert - 1.
\end{split}
\end{equation}

The inverse of the tangent function to approximate the Heaviside function, the model-driven generator can be defined as:
\begin{equation} \label{eq:tangent}
   H_{\epsilon}(\phi(x,y)) = \frac{1}{2}\left(1+\frac{2}{\pi}arctan\left(\frac{\phi(x,y)}{\epsilon}\right)\right). 
\end{equation}

\subsection{Derivatives w.r.t Shape Parameters }
Since $M=H_{\epsilon}(\phi(x,y))$, the derivatives of $M$ with respect to (w.r.t.) the parameters of a geometric shape can be transformed to those of $\phi$. The derivative of $M_{x,y}$ w.r.t. the parameter $\epsilon$ can be calculated as follows
\begin{equation}\label{EqBaseTrans}
\frac{\partial M_{x,y}}{\partial \epsilon}= \frac{1}{\pi} \frac{1}{1+(\frac{\phi(x,y)}{\epsilon})^2} \frac{-\phi(x,y)}{\epsilon^2}.
\end{equation} 
We take the parameter $a$ from detector outputs (i.e. $c_x, c_y, \theta, a, b$) as an example to introduce the gradient transfer of generator for updating detector parameters, the derivatives of parameter $a$, i.e. $\frac{\partial M_{x,y}}{\partial a}$,are calculated as follows 
\begin{equation}\label{EqBaseTrans}
\frac{\partial M_{x,y}}{\partial a}= \frac{1}{\pi} \frac{1}{1+(\frac{\phi(x,y)}{\epsilon})^2} \frac{\partial \phi(x,y)}{\partial a},
\end{equation}

For the shape of \textbf{Rotated ellipse}, the derivative $\frac{\partial \phi(x,y)}{\partial a}$ is easily to calculate as follows
\begin{equation}\label{EqDeriveRotaEllip}
\frac{\partial \phi(x,y)}{\partial a}= -\frac{2((x-c_x) \cos \theta+(y-c_y) \sin \theta)^2}{a^3},
\end{equation}
while $\frac{\partial M_{x,y}}{\partial c_x}$, $\frac{\partial M_{x,y}}{\partial c_y}$, $\frac{\partial M_{x,y}}{\partial b}$ and $\frac{\partial M_{x,y}}{\partial \theta}$ are derived similarly as $\frac{\partial M_{x,y}}{\partial a}$.

For the shape of \textbf{Rectangle}, we denote $\alpha=\alpha(c_x,a)\doteq \frac{x-c_x}{a}$ and $\beta=\beta(c_y,b)\doteq \frac{y-c_y}{b}$.
To obtain the derivative of $\phi(x,y)$ w.r.t. the four parameters, i.e. $w$,$h$,$c_x$,$c_y$, in Eq. \eqref{math:rectangle}, then the derivatives of $\phi$ w.r.t. the four parameters can be transformed those w.r.t. $\alpha$ and $\beta$ as follows
\begin{equation}\label{EqTransform}
\frac{\partial \phi}{\partial a}= \frac{\partial \phi}{\partial \alpha} \frac{\partial \alpha}{\partial a},\\
\end{equation}
where the terms alike $\frac{\partial \alpha}{\partial a}$ are easy to derive.
While the sub-gradient of $|x|$ w.r.t. $x$ is zero at the point $x=0$,
the derivative of $\frac{\partial \phi}{\partial \alpha}$ is obtained as follows
\begin{equation}\label{EqPartDeriv}
\frac{\partial \phi}{\partial \alpha}=
\left\{\begin{matrix}
\begin{array}{ll}
2 & \text{if } \alpha > |\beta|, \\
1 & \text{if } \alpha = |\beta|>0,\\
0 & \text{if } |\alpha| < |\beta| \text{ or } \alpha=\beta=0. \\
-1 & \text{if } \alpha = -|\beta|<0, \\
-2 & \text{if } \alpha < -|\beta|,\\
\end{array}
\end{matrix}\right.
\end{equation}
the derivative of $\frac{\partial \phi}{\partial \beta}$ can be similarly obtained.

For the shape of \textbf{Rotated Rectangle}, we denote $\alpha=\alpha(c_x,c_y,a,\theta)\doteq \\ \frac{(x-c_{x})cos\theta - (y-c_{y})sin\theta}{a}$ and $\beta=\beta(c_y,c_y,b,\theta)\doteq \frac{(x - c_{x})sin\theta + (y-c_{y})cos\theta}{b}$. The similar derivatives as Eq. \eqref{EqTransform} are derived as follows 
\begin{equation}\label{EqRotTransform}
\left\{\begin{matrix}
\begin{array}{l}
\frac{\partial \phi}{\partial a}= \frac{\partial \phi}{\partial \alpha} \frac{\partial \alpha}{\partial a},\\
\frac{\partial \phi}{\partial c_x}= \frac{\partial \phi}{\partial \alpha} \frac{\partial \alpha}{\partial c_x}+\frac{\partial \phi}{\partial \beta} \frac{\partial \beta}{\partial c_x}, \\
\frac{\partial \phi}{\partial \theta}= \frac{\partial \phi}{\partial \alpha} \frac{\partial \alpha}{\partial \theta}+\frac{\partial \phi}{\partial \beta} \frac{\partial \beta}{\partial \theta}. \\
\end{array}
\end{matrix}\right.
\end{equation}
where the derivative of $\frac{\partial \phi}{\partial \alpha}$ is the same as that in Eq. \eqref{EqPartDeriv}.

%
%
\bibliographystyle{splncs04}

\end{document}